\documentclass[conference]{IEEEtran}
\IEEEoverridecommandlockouts
\usepackage{cite}
\usepackage{amsmath,amssymb,amsfonts}
\usepackage{graphicx}
\usepackage{caption}
\usepackage{cuted}
\usepackage{algorithm}
\usepackage{algpseudocode}
\usepackage{textcomp}
\usepackage{xcolor}
\usepackage{amssymb}
\usepackage{amsmath}
\usepackage{hyperref}

\begin{document}

\title{Refine Where It Matters: Contact-Adaptive Meshing for Deformable Object Manipulation
}


\author{\IEEEauthorblockN{Gabriel Johnson}
\IEEEauthorblockA{\textit{The University of Tennessee, Knoxville} \\
 Knoxville, Tennessee \\
 gjohns57@vols.utk.edu}
\and
\IEEEauthorblockN{Tom Olesch}
\IEEEauthorblockA{\textit{The University of Tennessee, Knoxville} \\
 Knoxville, Tennessee \\
 tolesch@vols.utk.edu}
\and
\IEEEauthorblockN{Fei Liu}
\IEEEauthorblockA{\textit{The University of Tennessee, Knoxville} \\
 Knoxville, Tennessee \\
 fliu33@utk.edu
}}

\maketitle

\begin{abstract}
Deformable-object manipulation requires accurate modeling of highly localized contact while maintaining simulation speeds suitable for interactive use. Uniformly fine tetrahedral meshes can resolve these interactions, but they introduce unnecessary degrees of freedom throughout the object. Coarse meshes are more efficient, yet they may under-resolve contact-induced deformation. We present a contact-adaptive mesh refinement framework that introduces spatial resolution only where it is most useful. The method uses geometric, elastic, and contact quantities already available from the deformable-body solver to score candidate edges. A fully GPU-parallel procedure then selects a conflict-free subset and updates both mesh topology and physical state directly within the simulation loop. We evaluate the approach on real deformable-object manipulation sequences. Adaptive refinement reduces simulation-to-real error, improves agreement with a fine-mesh reference, and remains substantially more efficient than uniformly fine simulation. These results show that local refinement can improve physical fidelity without incurring the cost of a globally fine discretization. The framework provides a practical basis for contact-rich simulation in interactive manipulation, sim-to-real modeling, and online physical digital twins.
\end{abstract}

\section{Introduction}

Accurate and efficient modeling of deformable-object interaction is important for a broad range of robotic systems, including manipulation, surgical robotics, simulation-based planning, and interactive physical modeling \cite{Jihong_2022_RAM}. In these settings, a simulator must capture localized contact and deformation while remaining sufficiently fast to support repeated prediction, control, or model update. This requirement creates a persistent tension between physical fidelity and computational cost: increasing spatial resolution improves the ability to represent localized deformation, but also enlarges the system that must be assembled and solved at every simulation step.

\begin{figure}[!ht]
    \centering
    \includegraphics[width=1.0\linewidth]{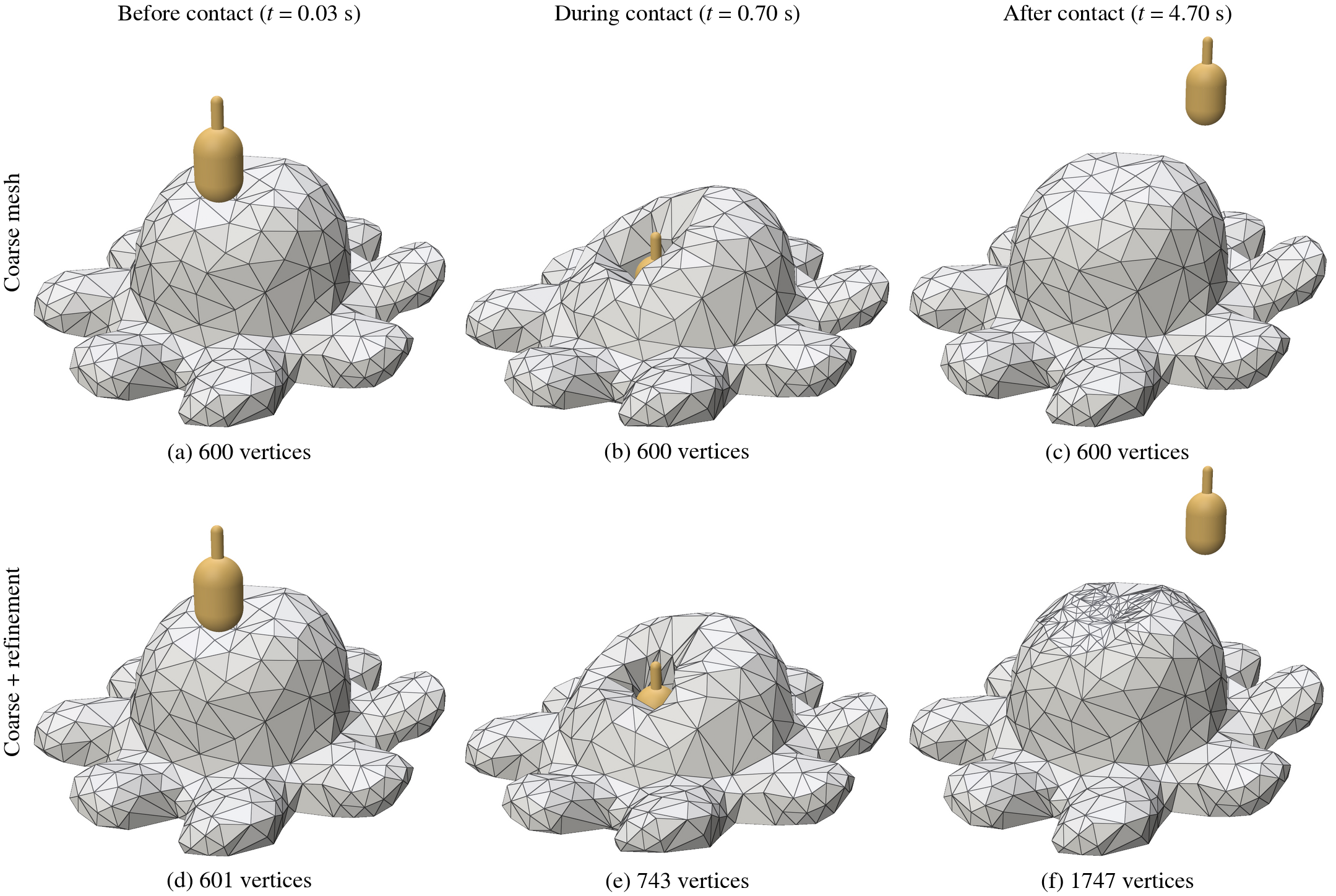}
    \caption{Contact-adaptive refinement during deformable-object manipulation. Starting from a coarse tetrahedral discretization, the mesh is refined online around the evolving tool-object interaction, introducing additional spatial resolution only where it is needed.}
    \label{fig:cover_figure}
\end{figure}

This challenge is especially pronounced for contact-rich manipulation \cite{Mark_2025_RAL, Jie_Tan_2023_pmlr-v229-zhang23e, liang2026isaacipccouplinghighfidelitysimulation}. The regions that require the highest spatial resolution are typically localized around the tool-object interaction and evolve continuously as contact moves across the deformable body. A fixed discretization must therefore either remain coarse and risk under-resolving these interactions, or be made uniformly fine and pay the corresponding computational cost everywhere. Figure~\ref{fig:cover_figure} illustrates the central idea explored in this work: rather than committing to a uniformly fine discretization, we adapt the mesh online so that resolution is introduced only where the current interaction requires it.

\subsection{Why Adaptive Refinement is Needed}

Deformable-object manipulation is characterized by strongly nonuniform spatial deformation. Large portions of an object may undergo smooth motion that can be represented adequately with a relatively coarse discretization, while tool contact can produce highly localized deformation, sharp geometric variation, and concentrated mechanical response near the interaction region~\cite{ferguson2023remeshing}. Representing these localized effects may require substantially higher spatial resolution than the surrounding object.

A fixed mesh therefore imposes a fundamental trade-off between computational efficiency and local fidelity. A coarse mesh limits the number of degrees of freedom and enables faster simulation, but may lack sufficient local resolution to represent contact-induced deformation. A uniformly fine mesh alleviates this limitation at the cost of introducing additional degrees of freedom throughout the entire object, including regions where the additional resolution provides little benefit. Similar trade-offs motivate reduced and adaptive representations for contact-rich deformable simulation~\cite{trusty2024trading}.

This limitation becomes particularly restrictive in robotic manipulation because contact is both localized and time-varying. Regions requiring high spatial resolution move with the tool and generally cannot be prescribed in advance. Pre-refining a fixed region is therefore insufficient for general interactions, while globally refining every potentially contactable region largely eliminates the computational advantage of a coarse discretization. Dynamic refinement instead provides a mechanism for allocating spatial resolution according to the evolving physical interaction~\cite{wicke2010dynamic,ferguson2023remeshing}.

The ability to selectively allocate simulation resolution is also important beyond a single forward simulation. Contact-aware manipulation requires sufficiently detailed local dynamics to predict the consequences of tool actions~\cite{Zilin_ICLR2024_1cc70be9}; simulation-to-real modeling requires those predictions to remain consistent with observed physical behavior~\cite{du2025embedded}; and online physical digital twins require models that can be updated repeatedly under tight computational budgets. More broadly, physics-based generative simulation and action-conditioned world models may require repeated rollouts of contact-rich dynamics, making uniformly high-resolution simulation increasingly expensive.

We therefore seek an adaptive discretization that begins from a coarse tetrahedral mesh and introduces additional degrees of freedom only where they are currently useful. By concentrating resolution around evolving contact and deformation while leaving the remainder of the object coarse, adaptive refinement provides a mechanism for balancing local physical fidelity with the computational efficiency required for interactive deformable-object modeling.

\subsection{Related Work}

\subsubsection{Adaptive Mesh Refinement}

Dynamic remeshing has been widely studied for physical simulation. Early work primarily focused on topology-changing phenomena such as fracture, tearing, and large inelastic deformation. Wicke et al.~\cite{wicke2010dynamic}, for example, introduced dynamic local remeshing for elastoplastic simulation with shear hardening. Adaptive remeshing has also been used in cloth simulation to introduce resolution around localized geometric features such as folds, wrinkles, and buckling.

More closely related to our setting, Ferguson et al.~\cite{ferguson2023remeshing} introduced in-timestep remeshing for tetrahedral elastodynamics and demonstrated that adaptive refinement and coarsening can improve the representation of contact. Their method provides a general and sophisticated remeshing framework, but is designed around a sequential CPU execution model. This makes direct application challenging in settings where topology adaptation must be performed repeatedly within an interactive simulation loop.

A complementary direction focuses on accelerating contact simulation without changing the underlying mesh topology. GIPC~\cite{Kemeng_Huang_2024_GIPC}, for example, develops a fully GPU-based formulation of incremental potential contact~\cite{Li2020IPC} and achieves efficient, robust contacting elastodynamics. Such methods substantially accelerate the physical solve, but the spatial resolution remains determined by the initial discretization. Consequently, accurately resolving localized contact still requires the relevant region to contain sufficient degrees of freedom a priori.

\subsubsection{Adaptive Resolution and Subspace Simulation}

Reduced-order and subspace methods address a related computational bottleneck by reducing the dimensionality of the system solved at each iteration. Du et al.~\cite{du2025embedded} apply reduced-space simulation to IPC for deformable-object manipulation, enabling efficient contact-rich simulation while retaining a fixed underlying full-space discretization. Trusty et al.~\cite{trusty2024trading} further introduce an adaptive subspace formulation in which the reduced basis is updated according to an oracle estimating solution quality.

These approaches adapt the space in which the dynamics are solved, whereas mesh refinement adapts the spatial discretization itself. A reduced basis can redistribute representational capacity within an existing full space, but it cannot introduce spatial degrees of freedom that are absent from the original mesh. This distinction is particularly relevant for sharp, spatially localized contact whose location is unknown before manipulation. Our work therefore complements reduced-order simulation by adapting the underlying mesh resolution directly around the evolving interaction.

\subsection{Our Contribution}

We develop a GPU-based adaptive refinement framework for deformable-object simulation that dynamically introduces additional degrees of freedom only where they are most useful. The method is designed for contact-rich settings in which the interaction region evolves over time, and is motivated by downstream applications such as \textit{contact-aware manipulation, sim-to-real modeling, online digital twins, and action-conditioned physical prediction}.

The main contributions of this work are:
\begin{itemize}
    \item We introduce a contact-adaptive tetrahedral refinement criterion that uses mesh geometry, elastic deformation, and contact response to determine where additional spatial resolution should be introduced.

    \item We develop a fully GPU-parallel refinement pipeline, including conflict-free edge selection, topology update, and field transfer, allowing the discretization to evolve directly within the simulation loop.

    \item We evaluate the method on real deformable-object manipulation sequences and show that local refinement reduces coarse-mesh discretization error around contact while remaining substantially more efficient than uniformly fine simulation.
\end{itemize}

\section{Approach}

\begin{figure}[!htb]
    \centering
    \includegraphics[width=1.0\linewidth]{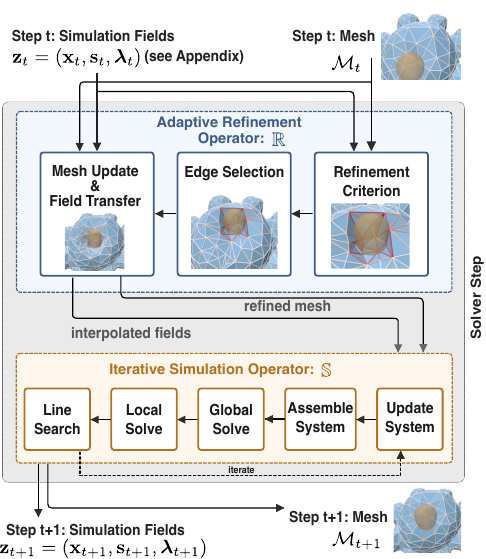}
    \caption{Our approach incorporates mesh refinement into the solver step.}
    \label{approach-figure}
\end{figure}

\subsection{Problem Definition}

We represent the deformable object at time step $t$ by a tetrahedral mesh
\begin{equation}
    \mathcal{M}_t=(\mathcal{V}_t,\mathcal{E}_t,\mathcal{T}_t),
\end{equation}
where $\mathcal{V}_t$, $\mathcal{E}_t$, and $\mathcal{T}_t$ denote the sets of vertices, edges, and tetrahedral elements, respectively. Our objective is to adapt the spatial discretization online by introducing additional degrees of freedom only in regions where finer resolution is expected to improve the representation of the current deformation and contact state. This setting naturally arises in the contact-rich manipulation sequences considered in our experiments, where the tool interacts with a localized portion of the deformable object and induces sharper spatial variation near the contact region than elsewhere in the mesh (see \ref{sec:experiments}). We therefore seek a refinement strategy that follows the evolving interaction region while keeping the remainder of the discretization coarse.


As illustrated in Fig.~\ref{approach-figure}, each solver step consists of two coupled stages: adaptive refinement and simulation iteration. Let
$\mathbf{z}_t=(\mathbf{x}_t,\mathbf{s}_t,\boldsymbol{\lambda}_t)$
denote the physical fields (i.e., the nodal positions, element-wise deformation variables, and Lagrange multipliers) associated with $\mathcal{M}_t$. The refinement operator $\mathbb{R}$ updates the mesh topology and transfers the current physical fields to the refined discretization, while the simulation operator $\mathbb{S}$ subsequently advances those fields on the updated mesh. Conceptually,
\begin{equation}
(\mathcal{M}_t,\mathbf{z}_t)
\xrightarrow{\;\mathbb{R}\;}
(\mathcal{M}_{t+1},\widetilde{\mathbf{z}}_t)
\xrightarrow{\;\mathbb{S}\;}
(\mathcal{M}_{t+1},\mathbf{z}_{t+1}),
\end{equation}
where $\widetilde{\mathbf{z}}_t$ denotes the physical fields transferred onto the refined mesh before the simulation iteration. The refinement operator $\mathbb{R}$ consists of the refinement criterion, conflict-free edge selection, mesh update, and field transfer in the following sections. The simulation operator $\mathbb{S}$ performs the subsequent finite-element solve on the refined mesh, including system assembly, global solve, line search, and local updates, as summarized in Algorithm~\ref{alg:refinement-step} and Appendix~\ref{app:simulation}.

The refinement operator evaluates the current mesh and physical state, selects and applies local edge splits, and transfers the associated fields to the refined discretization. The simulation operator then advances this refined state through the deformable-body solver to obtain the mesh and physical state at time step $t+1$.

Ideally, refinement would be driven directly by local discretization error. However, the underlying continuous deformation field is unavailable during simulation, so this error cannot be evaluated explicitly. We therefore infer the expected benefit of refinement from quantities available in the current physical state, including \textit{local mesh scale, elastic deformation, and contact response}. These quantities define an edge-wise refinement score, with larger values indicating edges for which additional spatial resolution is expected to be more beneficial.

The selected edges must also admit a valid local subdivision. Under the one-to-two tetrahedral splitting scheme used in this work, at most one edge of each tetrahedron may be selected,
\begin{equation}
    \sum_{e\subset T_k}
    \mathbf{1}[e\in\mathcal{S}_t]
    \leq 1,
    \qquad
    \forall\,T_k\in\mathcal{T}_t.
    \label{eq:compatibility_constraints}
\end{equation}
The refinement problem therefore consists of identifying high-value edges, selecting a mutually compatible subset, updating the affected mesh topology, and transferring the associated physical fields to the refined discretization so that computation can continue without re-initialization. The following sections describe the refinement criterion, conflict-free parallel edge selection, and mesh update with field transfer. The underlying deformable-body solver is summarized in the Appendix \ref{sec:appedix} \& \ref{app:contact}.

\begin{algorithm}[!tb]
\caption{Refinement in the simulation solver step}
\label{alg:refinement-step}
\begin{algorithmic}[1]
\Require $\mathbf{x}_t$, $\dot{\mathbf{x}}_t$, $\mathbf{s}_t$, $\boldsymbol\lambda_t$, $\mathcal{M}_t$, $\Delta t$
\Ensure $\mathbf{x}_{t+1},\dot{\mathbf{x}}_{t+1},\mathbf{s}_{t+1},\boldsymbol\lambda_{t+1},\mathcal{T}_{t+1}$

\Statex \textbf{Adaptive refinement}
\State \textit{Refinement criteria evaluation}
\State \textit{Candidate edge selection}
\State \textit{Mesh update \& field transfer} $\rightarrow \mathcal{M}_{t+1}$

\Statex \textbf{Newton iterations}
\For{$k = 1 \to K$}
    \State \textit{Update system} $\rightarrow H_\mathbf{x}, H_\mathbf{s}, G_\mathbf{x},G_\mathbf{s}, g_\mathbf{x}, g_\mathbf{s}, \mathbf{c}$
    $(\mathbf{x},\mathbf{s})$
    \State \textit{Assemble system} $\rightarrow H, g$
    $(\mathbf{H},\mathbf{g})$ in $\mathbf{x}$
    \State \textit{Global solve:} $\Delta\mathbf{x} \gets \mathbf{H}^{-1}(-\mathbf{g})$
    \State \textit{Local solve}
    \State \textit{Line search}
\EndFor

\State $\dot{\mathbf{x}}_{t+1} \gets (\mathbf{x}_{t+1}-\mathbf{x}_t)/\Delta t$
\end{algorithmic}
\end{algorithm}

\subsection{Contact-Adaptive Refinement Criterion}

The refinement criterion determines where additional degrees of freedom should be introduced before the next solver update. Rather than relying on a separate estimator, we construct this decision directly from quantities already available in the current deformable-body solve step from $t$ to $t+1$ shown in Fig. \ref{approach-figure}. In particular, the mesh geometry, elastic state, and contact response provide complementary indicators of where the existing discretization may be too coarse to represent the evolving interaction.

For each edge $e\in\mathcal{E}_t$, we define a refinement score
\begin{equation}
\label{eq:refinement_score}
r_e
=
r_e^{\mathrm{geo}}
+
r_e^{\mathrm{ela}}
+
r_e^{\mathrm{con}},
\end{equation}
where $r_e^{\mathrm{geo}}$, $r_e^{\mathrm{ela}}$, and
$r_e^{\mathrm{con}}$ measure the geometric, elastic, and contact-related benefit of refinement, respectively. Conceptually, $r_e$ estimates the utility of introducing a new degree of freedom along edge $e$: a high score indicates that the current physical state suggests finer spatial resolution would be useful in that region.

\paragraph{Geometric resolution}
The geometric term captures how coarse the current discretization is locally. For a piecewise-linear tetrahedral basis, larger elements provide fewer spatial degrees of freedom over the same physical region and are therefore more likely to under-resolve localized variation. We assign higher refinement priority to longer edges and normalize the edge contribution by the characteristic edge length of its incident tetrahedra. 
For an edge $e$ incident to tetrahedra $\mathcal{T}(e)$, we define
\begin{equation}
r_e^{\mathrm{geo}}
\propto
\frac{\ell_e}
{\sum_{T_k\in\mathcal{T}(e)} \ell(T_k)},
\end{equation}
where $\ell_e$ is the edge length and $\ell(T_k)$ denotes the total edge length of tetrahedron $T_k$.
This makes
$r_e^{\mathrm{geo}}$ reflect relative local mesh resolution rather than the absolute scale of the object.

\paragraph{Elastic deformation}
The elastic term uses the deformation state already evaluated by the finite-element solver. Each tetrahedron contributes an elastic energy through the constitutive model, and large elastic response indicates regions undergoing stronger deformation. Since the deformation within each tetrahedron is represented using the same low-order basis, strongly deformed regions can benefit from additional spatial degrees of freedom. For an edge $e$ incident to tetrahedra $\mathcal{T}(e)$,
\begin{equation}
r_e^{\mathrm{ela}}
\propto
\sum_{T_k\in\mathcal{T}(e)}
\Psi(\mathbf{s}_k),
\end{equation}
where $\Psi(\mathbf{s}_k)$ is the elastic energy density defined in the mixed FEM formulation in the Appendix \ref{app:simulation}). Thus, edges belonging to elements with stronger elastic response receive higher refinement priority.

\paragraph{Contact response}
The contact term couples refinement directly to the rigid-deformable interaction handled by the solver. The contact model evaluates the signed distance between the deformable surface and the rigid tool and converts this distance into a barrier response, as described in the Appendix. These quantities encode both proximity to contact and the strength of the local contact response. We define
\begin{equation}
r_e^{\mathrm{con}}
\propto
b(d_e),
\end{equation}
where $b(d)$ is the contact barrier energy defined in the Appendix. Consequently, edges closer to active contact receive larger refinement scores, while edges outside the contact region receive little or no contact contribution.

If an edge intersects the rigid surface, refinement is forced. In this case, inserting a new vertex introduces an additional degree of freedom inside the under-resolved contact region, allowing subsequent solver iterations to represent the local deformation and non-penetration response with greater spatial flexibility.

Edges whose refinement scores exceed a prescribed threshold form the candidate set for the subsequent conflict-free selection stage. The refinement score therefore provides a direct interface between the physical solve and the spatial discretization: \textit{the current solver state identifies regions where additional resolution is useful, local edge splits introduce new degrees of freedom in those regions, and the refined mesh is passed directly to the next finite-element solve.} In this way, the discretization adapts to the evolving deformation and contact rather than remaining fixed throughout the interaction.


\subsection{Parallel Conflict-Free Edge Selection}

The refinement criterion produces a set of high-value candidate edges, but these edges cannot be split independently. Under the one-to-two tetrahedral subdivision used in this work, each tetrahedron may contain at most one selected split edge. We therefore seek a conflict-free subset
$\mathcal{S}_t\subseteq\mathcal{C}_t$
that preserves high refinement scores while satisfying the compatibility constraint introduced in Eq. \ref{eq:compatibility_constraints}.

We perform this selection entirely on the GPU using a sequence of parallel kernels with one thread per tetrahedron. Shared edges are represented through a device-side hash map whose keys are mesh edges and whose values store their refinement scores and selection states. This provides a common reference for edges shared by multiple tetrahedra and avoids an explicit edge de-duplication stage.

\begin{figure}
    \centering
    \includegraphics[width=0.45\textwidth]{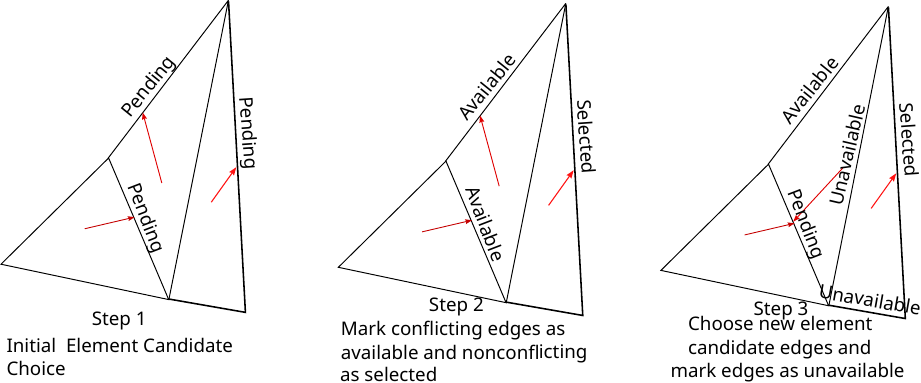}
    \caption{Nonconflicting candidate selection}
    \label{candidate-selection}
    \vspace{-0.5cm}
\end{figure}

\textit{Step 1}: As illustrated in Fig.~\ref{candidate-selection}, the selection proceeds iteratively. Each edge is assigned one of four states:
\textsc{Available}, \textsc{Pending}, \textsc{Selected}, or \textsc{Unavailable}.
Initially, all candidate edges are marked \textsc{Available}. Each tetrahedron then proposes its highest-scoring available edge, and all proposed edges are marked \textsc{Pending}.

\textit{Step 2}: Because neighboring tetrahedra may propose different edges that belong to the same tetrahedron, these proposals can violate the one-edge-per-tetrahedron constraint (Eq. \ref{eq:compatibility_constraints}). To resolve such conflicts, each tetrahedron examines its incident \textsc{Pending} edges and retains only the highest-scoring one; all other pending edges are returned to \textsc{Available}. An edge that remains \textsc{Pending} after this filtering is therefore the preferred available candidate for every tetrahedron incident to that edge. Such an edge is promoted to \textsc{Selected}.

\textit{Step 3}: Once an edge is selected, all other edges belonging to its incident tetrahedra are marked \textsc{Unavailable}, preventing them from being selected in later passes. Tetrahedra that remain unresolved then propose their next highest-scoring \textsc{Available} edge, and the process repeats for a fixed number of iterations.

This iterative filtering produces a mutually compatible set of refinement edges while preserving the local score ordering as much as possible. In our experiments, approximately five iterations were sufficient for most eligible tetrahedra to obtain a non-conflicting split edge. The resulting set $\mathcal{S}_t$ is then passed to the mesh-update stage for parallel edge splitting.

\subsection{Mesh Update and Field Transfer}

Given the conflict-free split set $\mathcal{S}_t$, we update the mesh topology by inserting a new vertex on each selected edge and subdividing its incident tetrahedra. The resulting discretization is
\begin{equation}
    \mathcal{M}_{t+1}
=
(\mathcal{V}_{t+1},\mathcal{E}_{t+1},\mathcal{T}_{t+1}),
\end{equation}
where $\mathcal{V}_{t+1}$ contains the newly introduced vertices and
$\mathcal{E}_{t+1}$ and $\mathcal{T}_{t+1}$ encode the updated edge and tetrahedral connectivity. Because the selected split edges are conflict-free, each affected tetrahedron undergoes a one-to-two subdivision and can be updated independently.

Changing the mesh topology also requires transferring the current physical state from $\mathcal{M}_t$ to $\mathcal{M}_{t+1}$. This ensures that refinement changes the spatial discretization without re-initializing the simulation. For nodal quantities, such as the deformed positions $\mathbf{x}$ defined in Appendix~\ref{app:simulation}, the state of a newly inserted vertex is obtained by linear interpolation from the endpoints of the split edge. For a midpoint split of edge $(i,j)$,
\begin{equation}
    \mathbf{x}_{\mathrm{new}}
    =
    \frac{1}{2}
    \left(
    \mathbf{x}_i+\mathbf{x}_j
    \right).
\end{equation}
Element-wise quantities are transferred directly from the parent tetrahedron. In the mixed FEM formulation, the deformation variables $\mathbf{s}_k$ and Lagrange multipliers $\boldsymbol{\lambda}_k$ are defined per element and are constant over each tetrahedron. The two child tetrahedra therefore inherit the corresponding element state from their parent. This preserves the current solver state while extending it to the refined discretization.

After topology update and field transfer, the refined mesh represents the same physical configuration with additional local degrees of freedom. The subsequent solver iterations can therefore continue directly on $\mathcal{M}_{t+1}$, allowing the newly introduced resolution to participate immediately in the deformation and contact solve shown in Fig \ref{approach-figure}.

\subsection{GPU-Parallel Implementation}

To minimize GPU execution overhead, we design the refinement and simulation pipeline to remain fully on-device and compatible with CUDA graph capture. Dynamic refinement, however, changes the number of active vertices, triangles, and tetrahedra over time, whereas captured kernel launches require fixed launch configurations. We address this mismatch by preallocating buffers for a maximum number of mesh primitives and maintaining device-side active counts. Kernels are launched with fixed dimensions, while threads whose indices exceed the current active count simply perform no operation. This allows the same captured execution graph to be reused as the mesh evolves without requiring host-side reconstruction or synchronization.

A second challenge arises from the sparse linear systems used by the deformable-body solver. Our implementation uses the block-sparse-row (BSR) operations provided by NVIDIA Warp~\cite{Macklin_Warp_A_High-performance_2022}. Constructing a BSR matrix requires fixed bounds on both the number of rows and the number of nonzero blocks per row, while mesh refinement dynamically changes the system sparsity pattern.

We obtain a fixed upper bound from the local tetrahedral connectivity. Each block row of the system Hessian corresponds to one vertex and contains nonzero blocks for vertices that share an incident tetrahedron. If a vertex is incident to at most $n_{\mathrm{tet}}^{\max}$ tetrahedra, each incident tetrahedron can introduce at most three neighboring vertices. The number of nonzero blocks in a row can therefore be bounded by
$
n_{\mathrm{nz}}^{\max}
=
3n_{\mathrm{tet}}^{\max}+1,
$
where the additional block corresponds to the diagonal. We preallocate the BSR structure according to this bound and update only the active entries as the topology changes. Together with fixed-capacity mesh buffers, this allows refinement, system assembly, and the subsequent solve to remain within the captured GPU execution pipeline.



\begin{figure}
    \centering
    \includegraphics[width=1.0\linewidth]{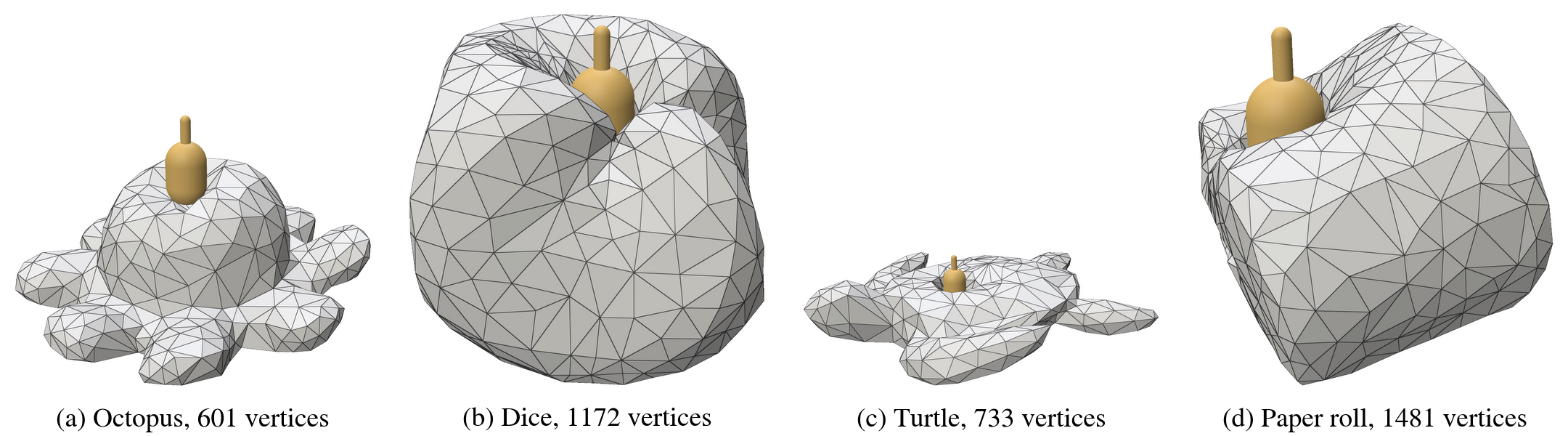}
    \caption{Refinement contact modeling on various models from the PokeFlex dataset.}
    \label{episodes-figure}
    \vspace{-0.5cm}
\end{figure}

\section{Experiments and Evaluation}
\label{sec:experiments}

We evaluated our model against real-world robot-soft-object interaction using the Pokeflex dataset from \cite{obrist2025pokeflex}. Pokeflex provides surface reconstructions for each frame in a collection of episodes with different deformable objects; however, from frame to frame, the surface meshes are not tracked meaning that there is not a clear connection between the vertices in one frame and those in the next. To address this, we use a private mesh tracking implementation built for the Pokeflex database which allows us to compare our simulated surface particle positions to the tracked particle positions producing a simulation to real loss metric.

\subsection{Experimental Setup}
 
In order to test the ability of the refinement solver to mitigate discretization error around sharp contacts, we built fine and coarse tetrahedral meshes using fTetWild \cite{10.1145/3386569.3392385} such that we could run the solver with high iteration counts on the fine mesh for a best-case, performance unconstrained baseline simulation. We then used the coarse meshes as samples of the ideal use case for this solver with few enough particles that the simulation can be run at interactive rates. 

Since our refinement solver has a number of parameters relating to how it selects when and where to refine, we first swept through a wide selection of parameter choices with the Pokeflex plush octopus episode to find a set of parameters that lead to the best possible error for the coarse meshes with refinement compared to the fine mesh.

Using these parameters, we computed the tracking error for the coarse mesh without refinement, the coarse mesh with refinement, and the fine mesh at each time step, as well as the error against the fine mesh simulation for the coarse mesh with and without refinement.

\subsection{Accuracy}

\begin{figure}[!t]
    \centering
    \includegraphics[width=0.45\textwidth]{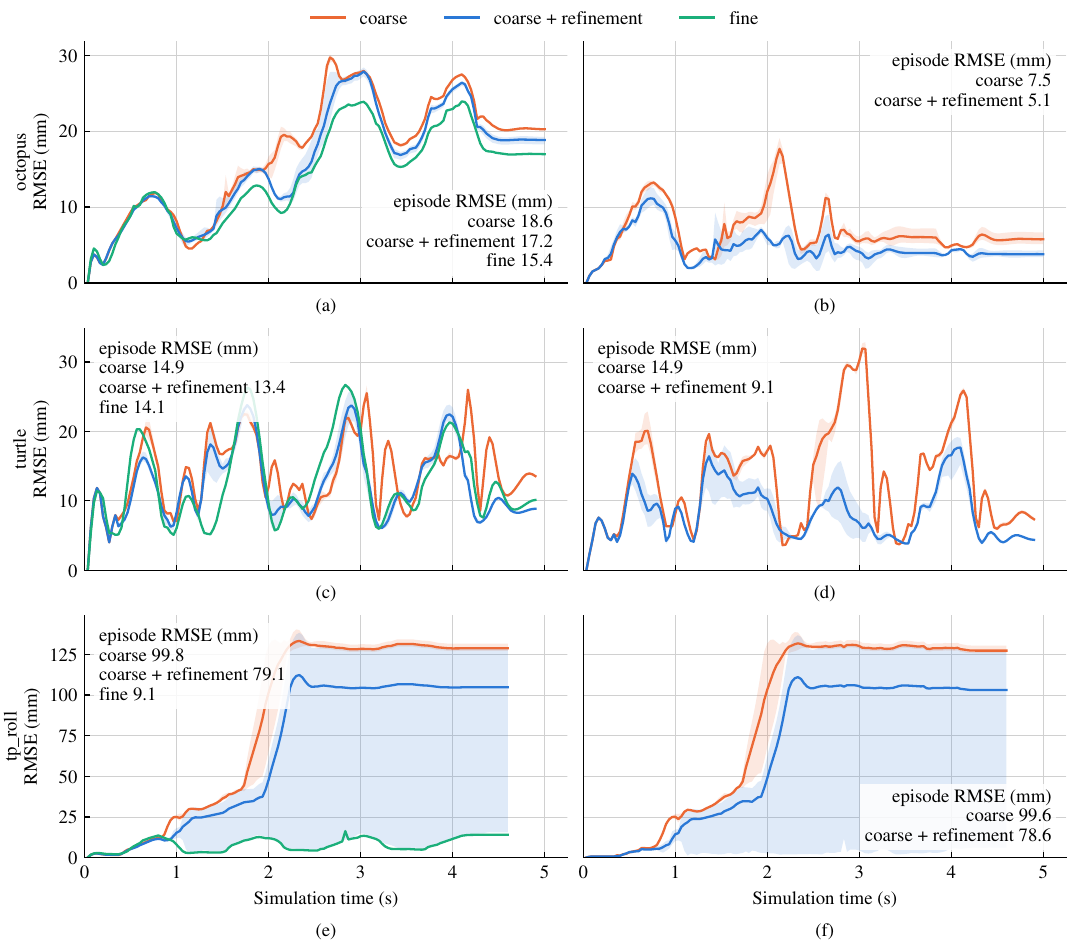}
    \caption{Comparison of errors against tracked mesh data (left) and comparison against fine mesh baseline (right). We see that while the tracking loss improvement of the refinement based solver as opposed to the base solver is minor, the refinement is significantly reducing discetization error when compared against the fine baseline. }
    \label{sim2real-comparison}
    \vspace{-0.5cm}
\end{figure}

\begin{figure}
    \centering
    \includegraphics[width=0.45\textwidth]{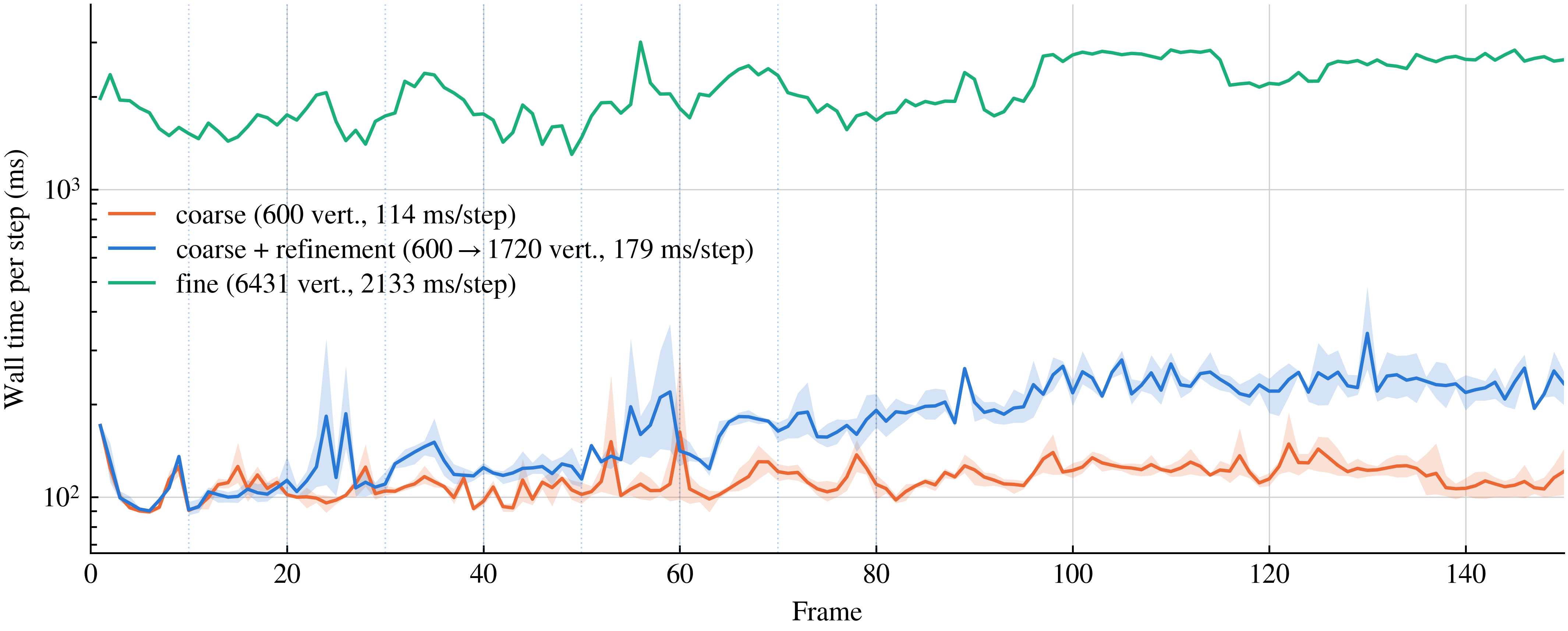}
    \caption{Timing comparison between coarse without refinement, coarse with refinement, and fine mesh simulation. The refinement solver slows down slightly as more vertices are added to the mesh but still remains about an order of magnitude faster than the fine simulation. All timings were performed on an NVidia RTX 4060 mobile GPU.}
    \label{sim-timings}
    \vspace{-0.5cm}
\end{figure}

Our experiments found that mesh refinement could provide a noticeable improvement in simulation loss, especially around sharp contacts. This improvement is somewhat hidden by the fact that the real world baseline cannot track the points under the tool tip accurately since they are obscured from the camera's view causing smoothing around the contact that reduces the potential simulation loss improvement we might have seen.

However, comparing the coarse mesh simulation with and without refinement to the fine mesh simulation, we see that in many cases we achieve much closer results to the fine mesh simulation with refinement than without it all while using many fewer vertices than the fine mesh. Figure \ref{sim2real-comparison} shows how the error between the coarse mesh and the fine mesh simulation spikes around times where there is deep contact, whereas the refinement simulation has significantly smaller error spikes during contact.

\subsection{Performance}

We were able to achieve simulation times fast enough to support interactive control of the tool-tip with real-time feedback on the deformation of the tissue while maintaining accuracy close to the fine mesh baseline. Figure \ref{sim-timings} shows the performance saving achieved by re-meshing only around the tool contact over using an overall fine mesh.

\begin{figure}
    \centering
    \includegraphics[width=0.45\textwidth]{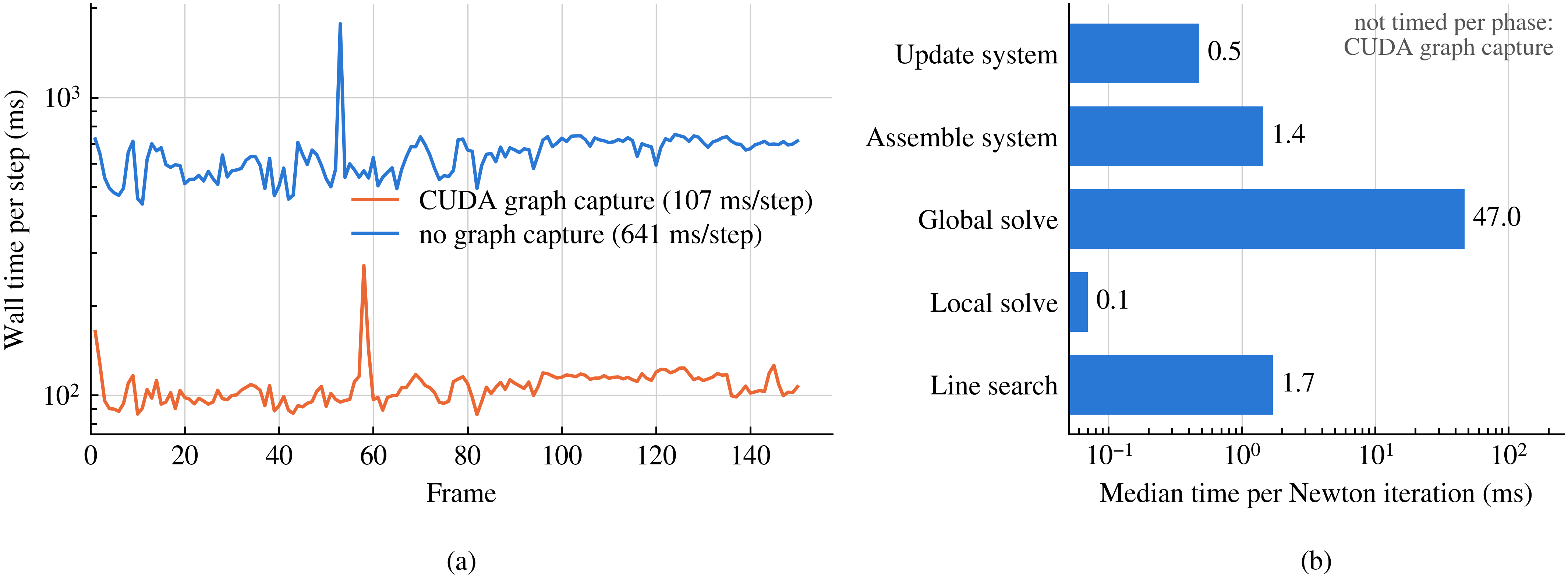}
    \caption{Comparison of step timings with and without CUDA graph capture (a) and sub-iteration timings (b).}
    \label{sim-timings-graph-gpu}
    \vspace{-0.5cm}
\end{figure}

Figure \ref{sim-timings-graph-gpu} shows that the iteration time is dominated by the global solve as is typical. One consequence of using the mixed FEM formulation from \cite{10.1145/3550469.3555418} is a slightly longer system assembly time since forming the system hessian in MFEM requires multiple expensive matrix multiplications. On the left, figure \ref{sim-timings-graph-gpu} demonstrates the importance of using graph capture to avoid kernel launch overhead, showing a performance savings of over five times when the conjugate gradient solve is not captured. This saving is made possible by the way in which the buffers are preallocated to a maximum capacity and refinement is performed between a swapping pair of buffers that get reused every iteration.

\section{Conclusion and Future Work}

These results show that mesh refinement can be a useful tool for fast simulation of robot soft manipulation allowing for fine representation of contact areas while keeping the mesh as a whole small. Our approach used a fairly simple score to determine where to refine; future work should explore more sophisticated metrics to determine which edges should be split and also where along the edge the split should occur.

Our implementation also suffered from performance costs in the conjugate gradient solve caused by introducing significant padding into the block sparse row system hessian. The performance of this approach could be significantly improved if the conjugate gradient implementation could be adapted to skip over these empty blocks within the graph capture.


The largest shortcoming of this implementation is the lack of mesh coarsening, forcing the number of particles in the simulation to only increase over time. This is acceptable for short episodes, but for longer episodes, the particle maximums would eventually be reached preventing any further refinement. Ferguson et al.'s approach \cite{ferguson2023remeshing} handles coarsening but adapting similar functionality to a GPU based solver was outside the scope of this project.

\appendix
\label{sec:appedix}
\section{Simulation Formulation}
\label{app:simulation}

\subsection{Mixed Variational FEM}

We use the mixed variational finite-element formulation of Trusty et al.~\cite{10.1145/3550469.3555418} as the underlying deformable-body solver. Let $\mathbf{X}$ denote material coordinates in the undeformed domain $\Omega$, and let $\mathbf{x}$ denote the corresponding deformed nodal positions. For each tetrahedral element $T_k\in\mathcal{T}$, the deformation gradient is denoted by $\mathbf{F}_k(\mathbf{x})\in\mathbb{R}^{3\times 3}$. The mixed formulation introduces an element-wise symmetric deformation variable $\mathbf{S}_k$, represented in vector form as $\mathbf{s}_k\in\mathbb{R}^6$, together with a Lagrange multiplier $\boldsymbol{\lambda}_k\in\mathbb{R}^6$ enforcing consistency between $\mathbf{S}_k$ and the symmetric component of $\mathbf{F}_k$.

Let $\mathbf{s}(\mathbf{F}_k)$ denote the vectorized symmetric component obtained from the polar decomposition of $\mathbf{F}_k$. The element consistency constraint is
\[
\mathbf{c}_k(\mathbf{x},\mathbf{s}_k)
=
\mathbf{D}\bigl(\mathbf{s}(\mathbf{F}_k(\mathbf{x}))-\mathbf{s}_k\bigr),
\]
where $\mathbf{D}=\operatorname{diag}(1,1,1,2,2,2)$ accounts for the duplicated off-diagonal entries of the symmetric tensor representation.

Using an elastic energy density $\Psi(\mathbf{s}_k)$, the discretized elastic potential is
\[
\widetilde{U}(\mathbf{x},\mathbf{s},\boldsymbol{\lambda})
=
\sum_{k\in\mathcal{T}}
\left[
\Psi(\mathbf{s}_k)
-
\boldsymbol{\lambda}_k^\top
\mathbf{c}_k(\mathbf{x},\mathbf{s}_k)
\right].
\]
Combining this potential with the inertial term gives the timestep objective
\[
\mathcal{L}(\mathbf{x},\mathbf{s},\boldsymbol{\lambda})
=
\frac{1}{2}
\|\mathbf{x}-\widetilde{\mathbf{x}}\|_{\mathbf{M}}^2
+
\Delta t^2
\widetilde{U}(\mathbf{x},\mathbf{s},\boldsymbol{\lambda}),
\]
where $\widetilde{\mathbf{x}}$ is the inertially predicted position and $\mathbf{M}$ is the lumped mass matrix.

We iteratively optimize the objective applying Sequential Quadratic Programming (SQP) to enforce satisfaction of the constraints. SQP is a variation of Newton-Raphson method applied to the Karush–Kuhn–Tucker (KKT) conditions necessarily satisfied at the minimum we seek. The KKT-conditions for our formulation are:
$$
    \begin{pmatrix}
        \nabla_\mathbf{x} f(\mathbf{x},\mathbf{s}) + G_\mathbf{x}^T \lambda \\
        \nabla_\mathbf{s} f(\mathbf{x},\mathbf{s}) + G_\mathbf{s}^t \lambda \\
        c(x,s)
    \end{pmatrix} = 0
$$
Here, $G_\mathbf{x}$ and $G_\mathbf{s}$ are the constraint Jacobians and $f(\mathbf{x}, \mathbf{s}) = ||\mathbf{x} - \Tilde{\mathbf{x}}||_M^2 +\Delta t^2 \Psi(\mathbf{s})$. In particular, $G_\mathbf{s}$ is a constant diagonal matrix, and $G_x$ does the work of encoding the mesh topology into the per iteration update linear system:
$$
\begin{pmatrix}
H_x & 0 & G_{\mathbf{x}}^T \\
0 & H_{\mathbf{s}} & G_{\mathbf{s}}^T \\
G_{\mathbf{x}} & G_{\mathbf{s}} & 0
\end{pmatrix}
\begin{pmatrix}
\delta \mathbf{x} \\
\delta \mathbf{s} \\
\lambda^{k+1}
\end{pmatrix}
=
\begin{pmatrix}
-g_\mathbf{x} \\
-g_\mathbf{s} \\
\mathbf{c}
\end{pmatrix}
$$
Where $H_\mathbf{s} = \nabla_{\mathbf{xx}}\mathcal{L}$ and $H_\mathbf{x}$ takes the place of $\nabla_{\mathbf{ss}}\mathcal{L}$ neglecting the second order derivatives of the constraints to avoid needing to compute a rotation hessian. On the right hand side, $g_\mathbf{x} = \nabla_\mathbf{x} f, g_s = \nabla_\mathbf{s} f$ for $f(\mathbf{x},\mathbf{s}) = ||\mathbf{x}-\Tilde{\mathbf{x}}||_M^2+\Psi(\mathbf{s})$. We can perform two Schur complements (block matrix Gauss-Jordan elimination) to transform this larger linear system into a symmetric positive definite system solving for the position degrees of freedom:

$$
\left(H_\mathbf{x} + G_\mathbf{x}^T H_\lambda G_\mathbf{x} \right) \delta \mathbf{x} = G_\mathbf{x}^T H_\lambda g_\lambda - g_\mathbf{x} \\
$$
where
$$
H_\lambda = G_\mathbf{s}^{-1}H_\mathbf{s}G_\mathbf{s}^{-1}
$$
$$
g_\lambda = \mathbf{c}(\mathbf{x}^{(t)}, \mathbf{s}^{(t)}) + G_\mathbf{s} H_\mathbf{s}^{-1}g_\mathbf{s}
$$

Using the position update, we apply two local systems for the Lagrange multiplier and deformation degree of freedom updates, respectively, that can be solved directly:

$$
\lambda^{(t + 1)} = -H_\lambda(g_\lambda - G_\mathbf{x}\delta\mathbf{x})
$$
$$
\delta\mathbf{s} = -H_\mathbf{s}^{-1}(G_\mathbf{s}\lambda^{(t + 1)} + g_\mathbf{s})
$$

$H_\mathbf{s}$ here is a block diagonal matrix consisting of the six-by-six element elastic energy hessians which may not be positive semi-definite for many useful elastic energies such as Neohookean so we apply a modified Cholesky factorization algorithm to project each element hessian to be PSD while producing an $LDL^T$ decomposition of each block allowing for quick computation of the inverted blocks for $H_\mathbf{s}^{-1}$.

We can efficiently take advantage of this systems sparsity and SPD nature by applying the conjugate gradient algorithm to local solve for the position update.

\subsection{Contact Model}
\label{app:contact}

\begin{figure}[!t]
    \centering
    \includegraphics[width=0.45\textwidth]{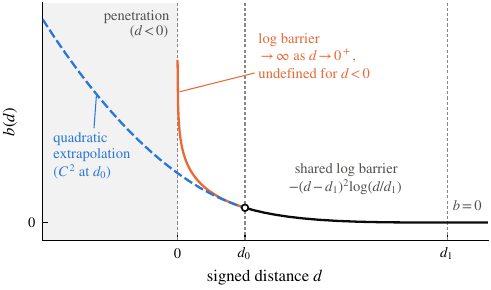}
    \caption{Comparison of quadratically extended log barrier versus standard log barrier}
    \label{barrier-fig}
    \vspace{-0.5cm}
\end{figure}

We augment the mixed FEM solver of Trusty et al. by incorporating a contact barrier similar to the one used in incremental potential contact (IPC) \cite{Li2020IPC}. IPC is very robust and provides a strong non-penetration guaranty; however, this guaranty comes with a significant amount of additional complexity in the form of the continuous collision detection line search and does not work well when the iteration count is cut short of full convergence. Since we are only interested in modeling rigid soft contact in this study, we opted to implement a simplified version of IPC without the continuous collision detection line search. Consequently, at any given iteration, non-penetration is not guaranteed so our version needs to handle resolving minor penetrations in a graceful way. We achieve this by modifying the original IPC barrier
$$
b_{\text{IPC}}(d) = \begin{cases}
    -(d - d_1)^2 \text{ln}\left(\frac{d}{d_1}\right) & \text{if } 0 < d \leq d_1\\
    0 & \text{otherwise}
\end{cases}
$$
by quadratically extrapolating the barrier near zero (see figure \ref{barrier-fig}) and using a signed distance function for $d$ instead of the unsigned one used in standard IPC. The resulting barrier is:
$$
b(d) = \begin{cases}
    b_\text{IPC}(d_0) + (d - d_0)b'_{\text{IPC}}(d_0) + \\  \frac{1}{2} (d-d_0)^2 b''_{IPC} (d_0) & \text{if } d \leq d_0 \\
    -(d - d_1)^2 \text{ln}\left(\frac{d}{d_1}\right) & \text{if } 0 < d < d_1 \\
    0 & \text{if } d_1 \leq d
\end{cases}
$$

Switching to this barrier energy has the two-fold advantage of avoiding numerical precision issues close to zero distance and allowing the barrier gradient to point in the direction of $-\nabla_x d$ even when $d$ is negative pushing any penetrated region back towards the nearest surface point of the rigid object.

Since this barrier hessian is not necessarily positive semi-definite we apply a simple Gauss-Newton approach dropping the term in the barrier hessian containing the second order derivative of the distance. More sophisticated techniques exist to achieve this PSD fix such as that proposed by Huang et al. \cite{Kemeng_Huang_2024_GIPC} however, this simpler fix met our needs with less implementation overhead.

Finally, to model real scenarios we need friction for which we use the lagged coulomb friction described in \cite{Li2021CIPC} where we fix normal forces at the beginning of each time step and use them to enforce a friction potential throughout that step.



\bibliographystyle{IEEEtran}
\bibliography{references}

\end{document}